\documentclass[runningheads]{llncs}
\usepackage[T1]{fontenc}
\usepackage{graphicx}
\usepackage{booktabs}
\usepackage{color}
\usepackage{booktabs}
\usepackage{multirow}
\usepackage{wrapfig}
\usepackage{adjustbox}
\usepackage{graphicx}
\usepackage{threeparttable} 

\usepackage{caption}
\usepackage[misc]{ifsym}
\newcommand{\corr}{(\Letter)}
\newcommand{\obs}{\mathbf{x}}
\newcommand{\view}{\mathbf{z}}
\newcommand{\InpS}{\mathcal{X}}
\newcommand{\OutS}{\mathcal{Y}}

\usepackage[utf8]{inputenc}
\usepackage{amsmath, amssymb, amsfonts}
\usepackage{tikz}
\usetikzlibrary{shapes.geometric, arrows.meta, positioning, fit, backgrounds, calc}
\usepackage{algorithm}
\usepackage{algorithmic}
\usepackage{subcaption}

\usepackage{dsfont}

\usepackage{mwe}
\usepackage{hyperref}
\begin{document}

\title{JEPAMatch: Geometric Representation Shaping for Semi-Supervised Learning}
\titlerunning{JEPAMatch}
\toctitle{JEPAMatch: Geometric Representation Shaping for Semi-Supervised Learning}

\author{Ali Aghababaei-Harandi\corr, Aude Sportisse \and Massih-Reza Amini}
\tocauthor{Ali Aghababaei-Harandi, Aude Sportisse, Massih-Reza Amini}

\institute{Université Grenoble Alpes, CNRS, \\Computer Science Laboratory LIG, Grenoble, France \\ \email{\{Firstname.Lastname}\}@univ-grenoble-alpes.fr}

\maketitle              

\begin{abstract}

Semi-supervised learning has emerged as a powerful paradigm for leveraging large amounts of unlabeled data to improve the performance of machine learning models when labeled data are scarce. Among existing approaches, methods derived from FixMatch have achieved state-of-the-art results in image classification by combining weak and strong data augmentations with confidence-based pseudo-labeling. Despite their strong empirical performance, these methods typically struggle with two critical bottlenecks: majority classes tend to dominate the learning process, which is amplified by incorrect pseudo-labels, leading to biased models. Furthermore, noisy early pseudo-labels prevent the model from forming clear decision boundaries, requiring prolonged training to learn informative representations. In this paper, we move beyond conventional logit-based output thresholding, toward an explicit shaping of geometric representations. Our approach is inspired by the recently proposed Latent-Euclidean Joint-Embedding Predictive Architectures (LeJEPA), a theoretically grounded framework asserting that meaningful representations should exhibit an isotropic Gaussian structure in latent space. Building on this principle, we propose a new training objective that combines adaptive pseudo-label selection with a latent-space regularization term.

Our proposed approach encourages well-structured representations while preserving the advantages of pseudo-labeling strategies. Through extensive experiments on CIFAR-100, STL-10 and Tiny-ImageNet, we demonstrate that the proposed method consistently outperforms existing baselines. In addition, our method significantly accelerates the convergence, drastically reducing the overall computational cost compared to standard FixMatch-based pipelines.

\end{abstract}

\section{Introduction}

Semi-supervised learning (SSL) \cite{books/mit/06/CSZ2006} has attracted increasing attention across a wide range of domains, including computer vision \cite{wang2022usb}, domain adaptation \cite{berthelot2021adamatch,da2022dual} and natural language processing \cite{chen2020mixtext}, as well as in application areas such as medicine \cite{eckardt2022semi} or biology \cite{moffat2021increasing}. Its appeal lies in its ability to leverage both labeled and unlabeled data during training, improving predictive performance or achieving comparable results with fewer labeled samples. This is particularly advantageous in real-world scenarios where acquiring large-scale labeled datasets is expensive and even impossible in some cases.

A variety of approaches have been explored in the literature, including pseudo-labeling methods \cite{lee2013pseudo,rizve2021defense}, consistency regularization techniques \cite{grandvalet2004semi,laine2017temporal,tarvainen2017mean}, and graph-based methods \cite{song2022graph}.  Recent deep learning SSL methods combine pseudo-labeling with consistency regularization and data augmentation, achieving remarkable empirical performance gains \cite{berthelot2019remixmatch,xie2020unsupervised,sohn2020fixmatch}. 

A particularly influential paradigm is FixMatch \cite{sohn2020fixmatch}, where predictions obtained from weakly augmented inputs supervise the model on strongly augmented versions of the same samples. Despite its empirical success, FixMatch suffers from two main limitations: pseudo-label selection relies on a fixed confidence threshold, which can bias training toward majority classes, and it requires a large number of iterations to converge. Adaptive or class-dependent thresholds have been proposed to mitigate the first issue \cite{xu2021dash,zhang2021flexmatch}, but the convergence speed problem remains unaddressed.

Recently, joint-embedding predictive architectures (JEPAs) \cite{assran2023self} have emerged as a promising alternative paradigm for representation learning in computer vision. JEPAs augment an encoder with a predictive model that directly predicts embeddings of augmented or local views, encouraging the encoder to capture semantically meaningful features without reconstructing raw inputs. 

Building on this idea, LeJEPA \cite{balestriero2025lejepa} proposes a theoretical framework suggesting that high-quality representations should follow an isotropic Gaussian distribution in the embedding space, characterized by equal variance along all directions and the absence of preferred axes. Overall, JEPA-based methods are primarily designed for stable representation learning; however, they may also promote more efficient feature learning and improved convergence dynamics.

In this work, we leverage the theoretically grounded architecture introduced in LeJEPA and integrate it into FlexMatch \cite{zhang2021flexmatch}, a variant of FixMatch with adaptive thresholding. Our proposed method, termed \textbf{JEPAMatch}, addresses the limitations of confidence-based semi-supervised learning by decoupling discrete classification training from the geometric organization of the representation space. 

To this end, the learning process is decomposed into two complementary levels: a Curriculum Level responsible for pseudo-label selection and a Representation Level that structures the learned feature space. Notably, JEPAMatch is not limited to FlexMatch and can be adapted to other FixMatch variants, such as FreeMatch and SoftMatch, further extending its applicability. By jointly optimizing these two levels, JEPAMatch improves classification performance, mitigates the class imbalance induced by confidence-based pseudo-labeling, and facilitates faster and more stable convergence. We empirically validate these advantages through extensive experiments on standard benchmark datasets,  CIFAR-100, STL-10 and Tiny-ImageNet.

\section{Related works}

Several variants of FixMatch \cite{sohn2020fixmatch} have been proposed to address limitations in pseudo-label selection. FlexMatch \cite{zhang2021flexmatch} introduces class-specific confidence thresholds to mitigate class imbalance. FreeMatch \cite{wang2022freematch} dynamically estimates both global and per-class thresholds. In addition, SoftMatch \cite{chen2023softmatch} replaces hard thresholding by weighting each unlabeled example according to its predicted confidence. Another line of work combines self-supervised learning with semi-supervised classification to improve feature representations and robustness. CRMatch~\cite{fan2023revisiting} adds a self-supervised rotation loss and an embedding distance loss. CoMatch~\cite{li2021comatch} and SimMatch~\cite{zheng2022simmatch} integrate contrastive learning and embedding alignment between weakly and strongly augmented samples. Finally, Suave~\cite{fini2023semi} leverages self-supervised clustering to generate more reliable pseudo-labels while improving representation quality.

However, these methods rely on contrastive or redundancy-reduction objectives that either require large batch sizes to maintain diverse negative pairs, as in SimCLR~\cite{chen2020simple}, or introduce cross-correlation whose computation scales quadratically with the embedding dimension, as in Barlow Twins~\cite{zbontar2021barlow}. In contrast, LeJEPA~\cite{balestriero2025lejepa} enforces representational isotropy via 1D random projections, requiring neither large batches, negative pairs, nor quadratic-cost statistics, making it particularly well-suited for integration into semi-supervised pipelines where computational efficiency and training stability are essential.

Semi-supervised methods often struggle with class imbalance and dimensional collapse, when the data are unbalanced and there are easier-to-learn classes. In this case, majority classes frequently exceed the confidence threshold $\tau$, leading to consistent gradient updates that further reinforce their dominance in the representation space, while minority classes suffer from sparse supervision. This effect results in poorly structured feature representations and ultimately degrades the model’s generalization performance.

In addition, curriculum-based SSL frameworks suffer from significantly slow convergence rates. Standard consistency regularization methods, such as FixMatch~\cite{sohn2020fixmatch} and its variants, typically require over  $2^{20}$ (one million) training iterations to converge. This bottleneck arises from the pseudo-label selection mechanism. At the outset of training, the classifier’s confidence is exceptionally low, so most unlabeled samples fail to surpass the confidence threshold, leaving the network with extremely sparse supervisory signals. Consequently, the model relies primarily on labeled data for many iterations, while most unlabeled samples —though processed— are repeatedly discarded, resulting in inefficient use of computational resources.

\section{Preliminaries: Semi-Supervised Learning Formulation}
In the standard semi-supervised learning (SSL) framework, we assume access to two datasets: a set of labeled training examples, denoted as
$\mathcal{X}_l = \{(\mathbf{x}_i, y_i)\}_{i=1}^n \in (\InpS \times \OutS)^n$, and a set of unlabeled training examples,
$\mathcal{X}_u = \{\mathbf{x}_{n+i}\}_{i=1}^u \in \InpS^u$.
Here, $\InpS \subseteq \mathbb{R}^d$ represents the input space, while $\OutS$ denotes the output space.

In this work, we focus on multiclass classification problems, where the output space is discrete: $\OutS = \{1, \dots, C\}$ with $C > 2$.
Both labeled and unlabeled samples are processed by a neural network architecture consisting of a feature extractor $f_\theta$, followed by a linear classifier $h_\phi$.
We further suppose that the supervised objective is defined using the standard cross-entropy loss applied to the labeled data.

\begin{equation}\label{eq:SSL_labeled_loss}
    \mathcal{L}_{\text{sup}} = \frac{1}{n} \sum_{i=1}^{n} H(y_i, p_i),
\end{equation}
where $H$ denotes the cross-entropy loss, defined as $H(y_i, p_i) = - \sum_{c=1}^{C} y_{i,c} \log(p_{i,c})$, with $y_{i,c}$ denoting the $c$-th component of the one-hot encoded label $y_i \in \{0,1\}^C$. Furthermore, $p_i = \text{softmax}(h_\phi(f_\theta(\obs_i)))$ represents the predicted class probabilities for the $i$-th labeled example, with \text{softmax} denoting the softmax function.

For unlabeled data, we adopt the consistency-based framework with data augmentations, as introduced in FixMatch~\cite{sohn2020fixmatch}. Specifically, each unlabeled sample $\obs_i$ is transformed into two distinct augmented views: a weakly augmented version $\obs_i^w$ and a strongly augmented version $\obs_i^s$. The weakly augmented view $\obs_i^w$ is first processed by the network to obtain the predicted class probabilities; $p_i^w = \text{softmax}(h_\phi(f_\theta(\obs_i^w)))$. A pseudo-label $\hat{y}_i^w$ is then generated from the model's prediction as $\hat{y}_i^w = \arg\max(p_i^w)$.

To ensure reliable supervision, we retain only predictions with sufficiently high confidence. A fixed confidence threshold $\tau$ is applied: if the maximum predicted probability satisfies $\max(p_i^w) > \tau$, the pseudo-label is deemed confident, and a binary mask $M_i = 1$ is assigned; otherwise, $M_i = 0$. The unlabeled objective enforces prediction consistency between the weak and strong views by applying the cross-entropy loss to the strongly augmented sample $\obs_i^s$, supervised by the pseudo-label $\hat{y}_i^w$ and gated by the confidence mask $M_i$:

\begin{equation}\label{eq:SSL_unlabeled_loss}
    \mathcal{L}_{\text{unsup}} = \frac{1}{u} \sum_{i=n+1}^{n+u} M_i H(\hat{y}_i^w, p_i^s),
\end{equation}
where $p_i^s = \text{softmax}(h_\phi(f_\theta(\obs_i^s)))$ denotes the predicted class probabilities for the strongly augmented view. The overall training objective integrates both the supervised loss~\eqref{eq:SSL_labeled_loss} and the unsupervised loss~\eqref{eq:SSL_unlabeled_loss} as follows:

\begin{equation}
    \mathcal{L}_{\text{total,ssl}} = \mathcal{L}_{\text{sup}} + \lambda_{\text{unsup}} \mathcal{L}_{\text{unsup}},
\end{equation}
where $\lambda_{\text{unsup}} > 0$ is a hyperparameter that controls the relative contribution of the unlabeled data to the total loss.

\section{Proposed method: JEPAMatch}

Our proposed method, JEPAMatch, mitigates the constraints of confidence-based semi-supervised learning by separating discrete classification training from the geometry of representation learning. It achieves this by breaking down the learning process into two synergistic parts: a Curriculum Level, which manages the selection of pseudo-labels, and a Representation Level, which defines the organization of the learned feature space.

The goal is to train a parameterized neural network comprising three key modules: a backbone encoder $f_\theta$, a classifier head $h_\phi$, and an additional projection head $g_\psi$. The encoder $f_\theta$ extracts latent representations from input features, the classifier $h_\phi$ generates class predictions, and the projection head $g_\psi$ learns geometry-aware representations to improve generalization to unseen data.

To capture both semantic and geometric features, JEPAMatch employs a multi-view augmentation strategy. For a given unlabeled image $\obs_i$, we generate three distinct types of views:

\begin{enumerate}
    \item \textbf{Weak View ($\obs_i^w$):} A single globally-sized crop with standard, mild augmentations (e.g., random flips or translations).
    \item \textbf{Strong View ($\obs_i^s$):} A single globally-sized crop subjected to heavy augmentations (e.g. color transformations or noise injection).
    \item \textbf{Local Views ($\{\obs_{i,k}^{loc}\}_{k=1}^K$):} $K$ smaller and local crops, mildly augmented to capture isolated parts of the image, each covering a small portion of the original image.
\end{enumerate}

\begin{figure*}[htbp]
    \centering
    \captionsetup[subfigure]{font=scriptsize}

    \begin{subfigure}[b]{0.12\textwidth}
        \centering
        \includegraphics[width=\linewidth]{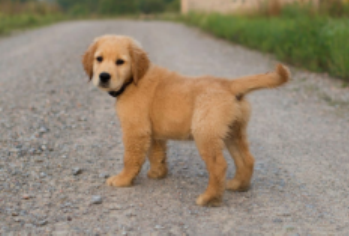}
        \caption{Original}
        \label{fig:aug_orig}
    \end{subfigure}%
    \hspace{1cm} 
    \begin{subfigure}[b]{0.12\textwidth}
        \centering
        \includegraphics[width=\linewidth]{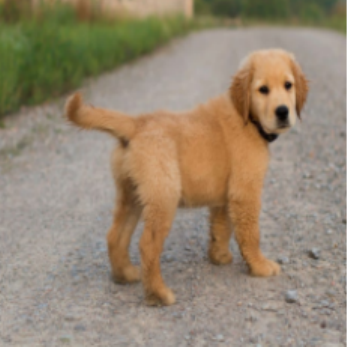}
        \caption{Weak}
        \label{fig:aug_weak}
    \end{subfigure}%
    \hspace{0.5cm} 
    \begin{subfigure}[b]{0.12\textwidth}
        \centering
        \includegraphics[width=\linewidth]{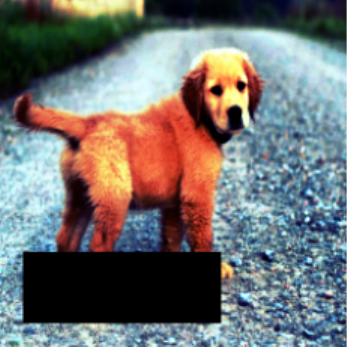}
        \caption{Strong}
        \label{fig:aug_strong}
    \end{subfigure}%
    \hspace{0.6cm} 
    \begin{subfigure}[b]{0.35\textwidth}
        \centering
        \includegraphics[width=0.2\linewidth]{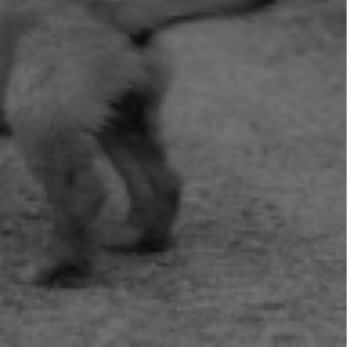} \hspace{0.2cm}
        \includegraphics[width=0.2\linewidth]{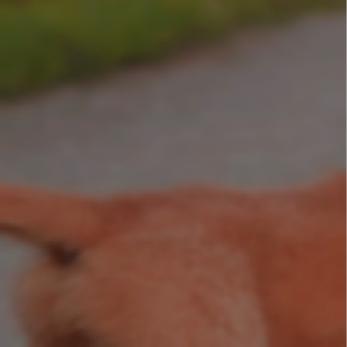} \hspace{0.2cm}
        \includegraphics[width=0.2\linewidth]{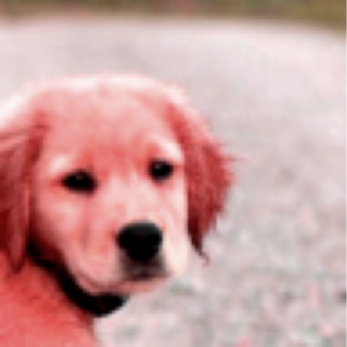}
        \caption{Local Views}
        \label{fig:aug_local}
    \end{subfigure}

    \caption{\footnotesize Multi-view Augmentation Strategy for JEPAMatch. The Original image is processed into global context views (Weak and Strong) and structurally isolated patches (Local views) to enforce geometric consistency.}
    \label{fig:augmentations}
\end{figure*}

These views are illustrated in Figure \ref{fig:augmentations}. The Curriculum Level strictly utilizes the global views ($\obs_i^w$ and $\obs_i^s$) to ensure the classifier has sufficient context to make accurate pseudo-label selection. Conversely, the Representation Level utilizes all views to enforce a robust, geometrically consistent latent space via self-supervised prediction across both global structures and local parts.

\subsection{Curriculum Level: Dynamic Pseudo-Labeling}
At the Curriculum Level, JEPAMatch is supervised using both ground-truth labels and high-confidence pseudo-labels generated by the classifier head $h_\phi$. For labeled data $\mathcal{X}_l$, we apply the standard supervised cross-entropy loss, as defined in Equation~\eqref{eq:SSL_labeled_loss}. For unlabeled data $\mathcal{X}_u$, we adopt the FlexMatch framework~\cite{zhang2021flexmatch}, which employs class-wise dynamic thresholds $\tau_c$. These thresholds, computed based on the number of samples assigned to each class, prevent majority-class dominance in the pool of confident pseudo-labels. Following the approach of FixMatch~\cite{sohn2020fixmatch}, prediction probabilities $p_i^w$ and their associated confident pseudo-labels $\hat{y}_i^w$ are generated for the weakly augmented view $\obs_i^w$. When a sample's confidence exceeds the dynamic threshold ($\max(p_i^w) \geq \tau_{\hat{y}_i^w}$), the mask $M_i$ is set to 1, and we apply the unsupervised cross-entropy loss to the strongly augmented view $\obs_i^s$, as described in Equation~\eqref{eq:SSL_unlabeled_loss}.

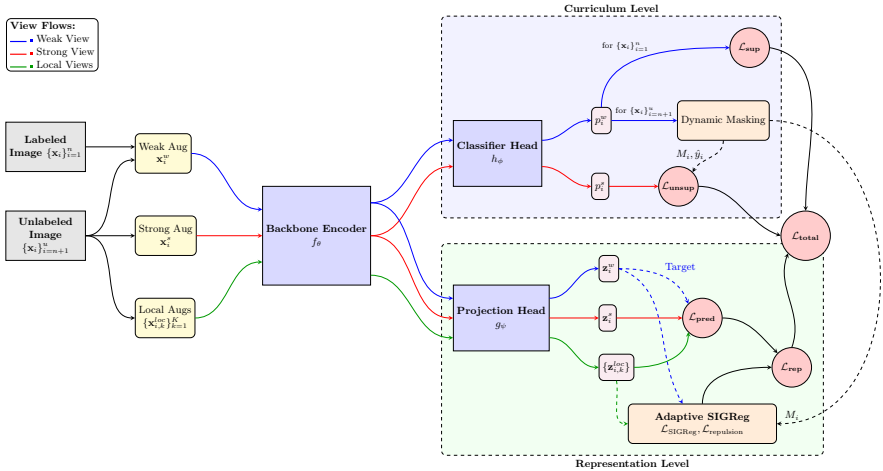
\begin{figure}[ht]
\centering
\resizebox{\textwidth}{!}{%
\begin{tikzpicture}[
    >=stealth,
    node distance=1.5cm and 2cm,
    data/.style={draw, cylinder, shape border rotate=90, aspect=0.25, minimum height=1.5cm, minimum width=1.2cm, fill=gray!10, align=center},
      image/.style={draw, rectangle, minimum height=1.5cm, text width=2.2cm, fill=gray!20, align=center, font=\bfseries},
    aug/.style={draw, rectangle, rounded corners, fill=yellow!20, minimum height=1.2cm, align=center, font=\small},
    model/.style={draw, rectangle, fill=blue!15, minimum height=3cm, minimum width=1.8cm, align=center, font=\bfseries},
    head/.style={draw, rectangle, fill=blue!15, minimum height=2cm, minimum width=1.8cm, align=center, font=\bfseries},
    feature/.style={draw, rectangle, rounded corners, fill=purple!8, minimum height=0.8cm, align=center, font=\small},
    loss/.style={draw, circle, fill=red!20, minimum size=1.2cm, align=center, font=\bfseries, inner sep=1pt},
    logic/.style={draw, rectangle, rounded corners, fill=orange!15, minimum height=1.2cm, align=center, font=\small},
    groupbox/.style={draw, dashed, thick, rounded corners, inner sep=10pt},
    arrow_w/.style={->, thick, blue},
    arrow_s/.style={->, thick, red},
    arrow_l/.style={->, thick, green!60!black},
    arrow_general/.style={->, thick, black},
    dashed_arrow_w/.style={->, dashed, thick, blue},
    dashed_arrow_s/.style={->, dashed, thick, red},
    dashed_arrow_l/.style={->, dashed, thick,green!60!black},
    arrow/.style={->, thick}
]

    \node[draw,  left=-0.1cm , yshift=0.8cm, rectangle, rounded corners, fill=white, align=left, font=\small] at (1.5, 5) {
        \textbf{View Flows:} \\
        \textcolor{blue}{\textbf{\rule{0.5cm}{1pt} \rule{2pt}{2pt}}} Weak View \\
        \textcolor{red}{\textbf{\rule{0.5cm}{1pt} \rule{2pt}{2pt}}} Strong View \\
        \textcolor{green!60!black}{\textbf{\rule{0.5cm}{1pt} \rule{2pt}{2pt}}} Local Views
    };

    \node[image] (input_u) {Unlabeled \\ Image $\{\obs_i\}_{i=n+1}^u$};
    
    \node[aug, right=1.5cm of input_u, yshift=2.5cm] (aug_w) {Weak Aug \\ $\obs_i^w$};

    \node[image, left=1.5cm of aug_w, yshift=0.2cm] (input_l) {Labeled \\ Image $\{\obs_i\}_{i=1}^n$};
        
    \node[aug, right=1.5cm of input_u, yshift=0cm] (aug_s) {Strong Aug \\ $\obs_i^s$};
    \node[aug, right=1.5cm of input_u, yshift=-2.5cm] (aug_l) {Local Augs \\ $\{\obs_{i,k}^{loc}\}_{k=1}^K$};

    \draw[arrow_general] (input_u.east) to[out=0, in=180] ([yshift=-0.2cm]aug_w.west);
    \draw[arrow_general] (input_u.east) to[out=0, in=180] (aug_s.west);
    \draw[arrow_general] (input_u.east) to[out=0, in=180] (aug_l.west);

    \draw[arrow_general] (input_l.east) to[out=0, in=180] ([yshift=0.2cm]aug_w.west);

    \node[model, right=2cm of aug_s] (encoder) {Backbone Encoder \\ $f_\theta$};

    \draw[arrow_w] (aug_w.east) to[out=0, in=180] ([yshift=0.8cm]encoder.west);
    \draw[arrow_s] (aug_s.east) to[out=0, in=180] (encoder.west);
    \draw[arrow_l] (aug_l.east) to[out=0, in=180] ([yshift=-0.8cm]encoder.west);

    \node[head, right=2.5cm of encoder, yshift=2.5cm] (classifier) {Classifier Head \\ $h_\phi$};
    
    \draw[arrow_w] ([yshift=1.0cm]encoder.east) to[out=0, in=180] ([yshift=0.4cm]classifier.west);
    \draw[arrow_s] ([yshift=0cm]encoder.east) to[out=0, in=180] ([yshift=-0.4cm]classifier.west);
    
    \node[feature, right=1.5cm of classifier, yshift=1cm] (prob_w) {$p_i^w$};
    \node[feature, right=1.5cm of classifier, yshift=-1cm] (prob_s) {$p_i^s$};
    
    \draw[arrow_w] ([yshift=0.4cm]classifier.east) to[out=0, in=180] (prob_w.west);
    \draw[arrow_s] ([yshift=-0.4cm]classifier.east) to[out=0, in=180] (prob_s.west);

    \node[logic, right=2cm of prob_w] (pseudo) {Dynamic Masking};
    \draw[arrow_w] (prob_w.east) -- (pseudo.west) node[pos=0.5, above, font=\scriptsize, text=black] {for $\{\obs_i\}_{i=n+1}^u$};

    \node[loss, above=1cm of pseudo, xshift=0.8cm] (loss_sup) {$\mathcal{L}_{\text{sup}}$};
    
    \draw[arrow_w] 
    (prob_w.north) to[out=90, in=180] 
    node[pos=0.55, above left, font=\scriptsize, text=black] {for $\{\obs_i\}_{i=1}^n$}
    (loss_sup.west);

    \node[loss, right=1.5cm of prob_s] (loss_u) {$\mathcal{L}_{\text{unsup}}$};
    \draw[arrow_s] (prob_s.east) -- (loss_u.west);
    
    \draw[->, dashed, thick] 
    (pseudo.south) to[out=270, in=60] 
    node[pos=0.4, right=-1.3cm, font=\small] {$M_i, \hat{y}_i$}
    (loss_u.north east);
    \node[head, right=2.5cm of encoder, yshift=-2.5cm] (projector) {Projection Head \\ $g_\psi$};

    \draw[arrow_w] ([yshift=1cm]encoder.east) to[out=0, in=180] ([yshift=0.6cm]projector.west);
    \draw[arrow_s] ([yshift=0cm]encoder.east) to[out=0, in=180] (projector.west);
    \draw[arrow_l] ([yshift=-1.2cm]encoder.east) to[out=0, in=180] ([yshift=-0.6cm]projector.west);

    \node[feature, right=1.5cm of projector, yshift=1.5cm] (z_w) {$\view_i^w$};
    \node[feature, right=1.5cm of projector, yshift=0cm] (z_s) {$\view_i^s$};
    \node[feature, right=1.5cm of projector, yshift=-1.5cm] (z_l) {$\{\view_{i,k}^{loc}\}$};

    \draw[arrow_w] ([yshift=0.6cm]projector.east) to[out=0, in=180] (z_w.west);
    \draw[arrow_s] (projector.east) to[out=0, in=180] (z_s.west);
    \draw[arrow_l] ([yshift=-0.6cm]projector.east) to[out=0, in=180] (z_l.west);

    \node[loss, right=2cm of z_s] (loss_pred) {$\mathcal{L}_{\text{pred}}$};
    \draw[arrow_s] (z_s.east) -- (loss_pred.west);
    \draw[arrow_l] (z_l.east) to[out=0, in=270] (loss_pred.south west);
    \draw[dashed_arrow_w] (z_w.east) to[out=0, in=110] node[midway, above right, font=\small, text=blue] {Target} (loss_pred.north west);

    \node[logic, below=2cm of loss_pred, minimum width=4.5cm] (sigreg) {
        \textbf{Adaptive SIGReg} \\
        $ \mathcal{L}_{\text{SIGReg}}, \mathcal{L}_{\text{repulsion}}$
    };
    
    \draw[->, dashed, thick, black]
        (pseudo.east) to[out=0, in=0, looseness=1.2]
        node[pos=0.99, above right, font=\small] {$M_i$}
        (sigreg.east);
        
    \draw[dashed_arrow_w] (z_w.east) to[out=340, in=120] ([xshift=-0.6cm, yshift=0.0cm]sigreg.north);
    \draw[dashed_arrow_l] (z_l.south) to[out=280, in=180] (sigreg.west);

    \node[loss, right=1.5cm of loss_pred, yshift=-1.5cm] (loss_rep) {$\mathcal{L}_{\text{rep}}$};
    \draw[arrow] (sigreg.north) to[out=90, in=180] (loss_rep.west);

    \node[loss, right=2.5cm of loss_u, yshift=-1.5cm, minimum size=1.5cm] (loss_tot) {$\mathcal{L}_{\text{total}}$};
    
    \draw[arrow] (loss_u.east) to[out=0, in=160] (loss_tot.west);
    \draw[arrow] (loss_sup.east) to[out=350, in=90] (loss_tot.north);
    \draw[arrow] (loss_pred.east) to[out=0, in=135] (loss_rep.north west);
    \draw[arrow] (loss_rep.north) to[out=70, in=250] (loss_tot.south west);

    \begin{scope}[on background layer]
        \node[groupbox, fill=blue!5, fit=(classifier) (prob_w) (prob_s) (pseudo) (loss_u) (loss_sup), label={[font=\bfseries, black]above:Curriculum Level}] {};
        \node[groupbox, fill=green!5, fit=(projector) (z_w) (z_l) (loss_pred) (sigreg) (loss_rep), label={[font=\bfseries, black!60!black]below:Representation Level}] {};
    \end{scope}

\end{tikzpicture}
}
\caption{Overview of the JEPAMatch architecture. The backbone encoder ($f_\theta$) processes distinct augmented views, with color-coded paths indicating the weak, strong, and local view flows. The \textbf{Curriculum Level} extracts global context to dynamically generate confident pseudo-labels ($\hat{y}_i$) and masks ($M_i$). The \textbf{Representation Level} aligns global and local views via JEPA prediction while applying Adaptive Class-wise SIGReg to tighten without dimensional collapse. Red circles denote loss terms, while the Dynamic Masking and Adaptive SIGReg modules compute the regularization, SIGReg and repulsion. Dashed lines indicate data passed as targets to subsequent modules. Notations are consistent with methods.}
\label{fig:jepamatch}
\end{figure}

The combination of labeled and unlabeled losses, as introduced in Equations~\eqref{eq:SSL_labeled_loss} and~\eqref{eq:SSL_unlabeled_loss}, serves as the primary discriminative objective. This objective encourages the latent features to form discrete, class-specific clusters, which is essential for achieving high classification accuracy. However, such clustering inherently risks geometric collapse of the latent space, an issue we mitigate through the isotropic constraints introduced at the Representation Level. Figure \ref{fig:jepamatch} depicts the JEPAMatch architecture, combining a backbone encoder with curriculum-based pseudo-labeling and adaptive class-wise SIGReg to align global/local features, mitigate class imbalance, and prevent representation collapse.

\subsection{Representation Level: Self-Supervised Geometry}
To ensure that the network learns robust and invariant features, the backbone outputs are passed through a projection head $g_\psi$ to obtain $\view = g_\psi(f_\theta(\obs))$. Accordingly, we denote the representations of the strong, weak, and local views as $\view_i^s$, $\view_i^w$, and $\{\view_{i,k}^{loc}\}_{k=1}^K$, respectively. At the Representation Level, three distinct losses are introduced, as detailed below.

\subsubsection{Self-Supervised Prediction Loss}
The core of our representation learning framework is the Joint Embedding Predictive Architecture (JEPA) loss~\cite{assran2023self}. This loss is motivated by the principle that the representation of a local part of an image—or its heavily augmented version—should align closely with the representation of the weakly augmented global image. This encourages the model to learn a geometrically meaningful representation of images. To achieve this, we treat the weak view representations $\view_i^w$ as targets, which the model predicts using the strongly augmented views $\view_i^s$ and the local crops $\{\view_{i,k}^{loc}\}_{k=1}^K$. The prediction loss is formalized as the distance between the predicted local/strong views and the target weak view:

\begin{equation}
    \mathcal{L}_{\text{pred}} = \frac{1}{u}\sum_{i=n+1}^{n+u}\left(\mathcal{D}(\view_i^s, \view_i^w) + \sum_{k=1}^K\mathcal{D}(\view_{i,k}^{loc}, \view_i^w)\right),
\end{equation}

\noindent
where $\mathcal{D}$ denotes a distance metric (e.g., Mean Square Error or Cosine distance). While $\mathcal{L}_{\text{pred}}$ effectively aligns local parts with the global context, it remains highly susceptible to dimensional collapse~\cite{balestriero2022contrastive}. To address this, we introduce a structural regularizer, detailed in the following section.

\subsubsection{Adaptive Class-wise SIGReg}
Inspired by the LeJEPA framework~\cite{balestriero2025lejepa}, which theoretically demonstrates that an isotropic Gaussian distribution is optimal for minimizing downstream prediction risk in latent embeddings, we extend the Sketched Isotropic Gaussian Regularization (SIGReg) mechanism. SIGReg prevents dimensional collapse by mapping high-dimensional features through 1D random projections (sketches) and matching their Empirical Characteristic Function to that of a standard normal distribution $\mathcal{N}(0, I)$. However, while global SIGReg is well-suited for generic representation learning, it conflicts with the discriminative objectives of semi-supervised classification, where data must form non-overlapping class clusters.

To reconcile these goals, we propose Adaptive Class-Wise SIGReg. Rather than enforcing a single global Gaussian distribution, we encourage each class to form its own isolated isotropic Gaussian, characterized by a class-specific mean $\mu_c$ and a tightly controlled variance $\sigma$. Our approach consists of two training phases, detailed in the following paragraphs.

\paragraph{Warmup Phase: Global SIGReg.} At the beginning of the training, the classifier's predictions are highly unstable, the pseudo-labels are largely unreliable. Attempting to build class-specific clusters at this stage would lead to severe confirmation bias, forcing the geometry to organize around incorrect predictions. Therefore, during the initial warmup phase (first $T_{\text{warm}}$ training iterations), we temporarily ignore class assignments and apply standard SIGReg loss as in LeJEPA \cite{balestriero2025lejepa} on the projection vectors $\{\view_{i,k}^{loc}\}_{k=1}^K$. By forcing the distribution of $\{\view_{i,k}^{loc}\}_{k=1}^K$ to be $\mathcal{N}(0, I)$, we ensure the network utilizes all available dimensions and builds a uncollapsed feature space while the prediction loss ($\mathcal{L}_{\text{pred}}$) learns basic structural invariance. In this phase, the representation loss is defined as:

\begin{equation}
    \mathcal{L}_{\text{rep}} = (1-\beta)\mathcal{L}_{\text{pred}} + \beta \frac{1}{K}\sum_{k=1}^K\text{SIGReg}(\{\view_{i,k}^{loc}\}_{i=n+1}^{n+u}, \mathcal{N}(0, I))
\end{equation}
where $\beta$ is a hyperparameter balancing the loss terms. We apply the SIGReg objective just to the local views, because the prediction loss ($\mathcal{L}_{\text{pred}}$) forces these local features to align with the global context, the weak and strong views implicitly inherit the desired Gaussian normality. 

\paragraph{Main Phase: Adaptive Class-Wise SIGReg.} Once the warmup phase concludes, the model has developed sufficient confidence to begin shaping the latent space. During this phase, the regularizer gradually shifts from structuring a continuous global manifold to forming tight class clusters.

\textbf{Class Mean:} 
For each class $c$, the target mean $\mu_c$ is computed from both the labeled data and the confident pseudo-labeled data present in the batch:
\begin{equation}
    \mu_c = \frac{1}{|B_c|} \sum_{\view^w_c \in B_c} \view^w_c
\end{equation}
where $B_c$ is the combined set of labeled and confident unlabeled samples for class $c$. To evaluate the normality of these newly forming clusters, we conditionally center the projection vectors $\view_i^{loc}$ based on the FlexMatch confidence mask $M_i$:
\begin{equation}
    \hat{\view}_{i,k}^{loc} = \view_{i,k}^{loc} - (M_i \cdot \mu_{\hat{y}_i})
\end{equation}
If a sample is confident ($M_i = 1$), it is perfectly centered by its class mean. This isolates the residual vector, allowing us to apply the normality constraint directly to the intra-class variance. If a sample is unconfident ($M_i = 0$), no class mean is subtracted, and the sample is regularized with respect to the global zero-mean SIGReg target until the model assigns it a reliable pseudo-label.

\textbf{Variance Annealing:} Because the confident data are now centered around specific class means, the geometric objective shifts from expanding the space to forming tight clusters. The target standard deviation $\sigma$ is annealed linearly from $1.0$ to $0.1$ during the main phase according to:
\begin{equation}\label{var:anneal}
    \sigma(t) = 1.0 - 0.9 \cdot \frac{t - T_{\text{warm}}}{T - T_{\text{warm}}}
\end{equation}
where $t$ is the current iteration, $T_{\text{warm}}$ is the number of warmup iterations, and $T$ is the total number of training iterations. For the centered projection vectors $\{\hat{\view}_{i,k}^{loc}\}_{k=1}^K$, we enforce a distribution $\mathcal{N}(0, \sigma^2 I)$.

\textbf{Active Repulsion Loss:} To prevent the network from trivially minimizing $\mathcal{L}_{SIGReg}$ by collapsing all class means into a single point, we introduce an Active Repulsion Loss. We extract mean $\mu_c$ of each valid class present in the batch. We normalize these means ($\bar{\mu}_c$) and create matrix $\bar{\mu}$ to compute their cosine similarity matrix $S = \bar{\mu} \bar{\mu}^T$. We penalize positive similarity between distinct classes:
\begin{equation}\label{eq:repulsion_loss}
    \mathcal{L}_{\text{repulsion}} = \frac{1}{C_\mu(C_\mu-1)} \sum_{i \neq j}^{C_\mu} \max(0, S_{i,j})^2,
\end{equation}
where $C_\mu$ denotes the number of unique class means active in the current iteration. If fewer than two classes are active, we set $\mathcal{L}_{\text{repulsion}}=0$. Representation loss $\mathcal{L}_{rep}$ for the main phase can be written as:
\begin{equation}
    \mathcal{L}_{\text{rep}} = (1-\beta)\mathcal{L}_{\text{pred}} + \beta \frac{1}{K}\sum_{k=1}^K\text{SIGReg}(\{\hat{\view}_{i,k}^{loc}\}_{i=n+1}^{n+u}, \mathcal{N}(0, \sigma^2I)) + \mathcal{L}_{\text{repulsion}}.
\end{equation}

\begin{algorithm}[t!]
\caption{JEPAMatch}
\label{alg:jepamatch}
\begin{algorithmic}[1]
\STATE \textbf{Input:} Labeled ($\mathcal{X}_l$) and Unlabeled ($\mathcal{X}_u$) sets, Threshold base ($\tau$), Number of total ($T$) and warmup iterations ($T_{warm}$), Loss hyper-parameters ($\lambda_{\text{unsup}}$, $\lambda_{\text{rep}}$, $\beta$).
\STATE \textbf{Initialize:} $f_\theta, h_\phi, g_\psi$ (Backbone, Classifier, Projection).
\FOR{$t = 1$ \TO $T$}
    \STATE Sample batch $B_l$ from $\mathcal{X}_l$ and $B_u$ from $\mathcal{X}_u$.
    \STATE Generate $\obs_i^w$ for each $\obs_i \in B_l$ and $\{\obs_i^w, \obs_i^s, \{\obs_{i,k}^{loc}\}_{k=1}^K\}$ for each $\obs_i \in B_u$
    \STATE Compute $\mathcal{L}_{\text{sup}} = \frac{1}{|B_l|} \sum_{i\in B_l} H(y_i, \text{softmax}(h_\phi(f_\theta(\obs_i^w)))$.
    \STATE Compute $p_i^w$ for $\obs_i^w$; derive pseudo-labels $\hat{y}_i^w$ and confidence $\text{conf}_i=\max(p_i^w)$.
    \STATE Update dynamic thresholds $\tau_{\hat{y}_i}$ and compute mask $M_i = \mathds{1}(\text{conf}_i \geq \tau_{\hat{y}_i})$.
    \STATE $\mathcal{L}_{\text{unsup}} = \frac{1}{|B_u|} \sum_{i\in B_u} M_i H(\hat{y}_i^w, \text{softmax}(h_\phi(f_\theta(\obs_i^s))))$.
    
    \STATE Compute $\view_i^w, \view_i^s, \{\view_{i,k}^{loc}\}_{k=1}^K$ via projection head $g_\psi$.
    \STATE $\mathcal{L}_{\text{pred}} = \frac{1}{|B_u|}\sum_{i\in B_u}\left(\mathcal{D}(\view_i^s, \view_i^w) + \sum_{k=1}^K\mathcal{D}(\view_{i,k}^{loc}, \view_i^w)\right)$
    
    \IF{$t < T_{warm}$}
        \STATE $\mathcal{L}_{\text{rep}} = (1-\beta)\mathcal{L}_{\text{pred}} + \beta \frac{1}{K}\sum_{k=1}^K\text{SIGReg}(\{\view_{i,k}^{loc}\}_{i=n+1}^{n+u}, \mathcal{N}(0, I))$
    \ELSE
        \STATE Compute batch class means: $\mu_c = \text{Mean}(\view_i^w \in B_c)$, where $B_c$ is the set of labeled and confident unlabeled samples for class $c$. \\
        \STATE Center features: $\hat{\view}_{i,k}^{loc} = \view_{i,k}^{loc} - (M_i \cdot \mu_{\hat{y}_i})$.
        \STATE Anneal $\sigma_t$ and compute $\mathcal{L}_{\text{repulsion}}$ given in \eqref{var:anneal} and \eqref{eq:repulsion_loss} between distinct $\mu_c$.
        \STATE $\mathcal{L}_{\text{rep}} = (1-\beta)\mathcal{L}_{\text{pred}} + \beta \frac{1}{K}\sum_{k=1}^K\text{SIGReg}(\{\hat{\view}_{i,k}^{loc}\}_{i=n+1}^{n+u}, \mathcal{N}(0, \sigma^2I)) + \mathcal{L}_{\text{repulsion}}$
    \ENDIF
    
    \STATE $\mathcal{L}_{\text{total}} = \mathcal{L}_{\text{sup}} + \lambda_{\text{unsup}} \mathcal{L}_{\text{unsup}} + \lambda_{\text{rep}} \mathcal{L}_{\text{rep}}$.
    \STATE Update $f_\theta, h_\phi, g_\psi$ via backpropagation.
\ENDFOR
\STATE \textbf{Return:} $f_\theta, h_\phi, g_\psi$
\end{algorithmic}
\end{algorithm}

\subsection{Total Loss Formulation}
JEPAMatch optimizes the network by jointly minimizing the curriculum prediction loss and the Representation loss. The total loss is formulated as:
\begin{equation}
    \mathcal{L}_{\text{total}} = \mathcal{L}_{\text{sup}} + \lambda_{\text{unsup}} \mathcal{L}_{\text{unsup}} + \lambda_{\text{rep}} \mathcal{L}_{\text{rep}}
\end{equation}
where the $\lambda$ hyper-parameters balance the geometric constraints against the predictive tasks. The representation loss, $\mathcal{L}_{\text{rep}}$, varies depending on the training phase. 

Our proposed approach is described in Algorithm~\ref{alg:jepamatch}: line 5 corresponds to data augmentation, lines 6-9 implement the Curriculum Level, lines 10-18 implement the Representation Level, line 13 computes the representation loss during the warmup phase, while lines 16-18 correspond to the main phase, and finally lines 20-21 compute the total loss and perform the standard parameter updates via backpropagation. In the next section, we validate our method through a series of experiments on different datasets.

\section{Experimental Setup}
\textbf{Datasets.} We conducted experiments on three challenging SSL benchmarks: CIFAR-100 \cite{krizhevsky2009learning} and STL-10 \cite{coates2011analysis}, and Tiny-ImageNet \cite{le2015tiny}. For CIFAR-100, which contains 50,000 images across 100 classes, we adopted the splitting scenarios proposed by FixMatch \cite{sohn2020fixmatch}, using 400, 2,500, and 10,000 labeled images with WRN-28-8 \cite{zagoruyko2016wide} as the backbone network. For STL-10, we used splits of 40 and 1,000 labeled images with WRN-37-2 \cite{zagoruyko2016wide} as the backbone. Finally, for Tiny-ImageNet, which consists of 100,000 images, we considered scenarios with 1,000 and 10,000 labeled images, using ResNet-50 \cite{he2016deep} as the backbone. In ablation experiments, we used WRN-28-2 as the backbone network. Weak and strong augmentations are consistent with those used in FlexMatch \cite{zhang2021flexmatch}. Across all experiments, we extracted \( K = 6 \) local crops per image, with crop areas sampled between 0.2 and 0.5 for CIFAR-100, and between 0.1 and 0.4 for STL-10 and Tiny-ImageNet. 
The projection head $g_\psi$ is implemented as a 3-layer  Multi-Layer Perceptron (MLP). To account for the increased complexity and number of classes across our benchmarks, the final output dimension is set to 128 for CIFAR-100 and STL-10, and increased to 256 for Tiny-ImageNet. Image classification results are reported as error rate (\%), defined as $100 - \text{accuracy}$ (\%).

\textbf{Baselines.} We compared JEPAMatch against state-of-the-art SSL approaches. This includes standard consistency regularization methods such as FixMatch \cite{sohn2020fixmatch}, AdaMatch \cite{berthelot2021adamatch}, and FlexMatch \cite{zhang2021flexmatch}, as well as recent advancements like SoftMatch \cite{chen2023softmatch}, FreeMatch \cite{wang2022freematch}, and RegMixMatch \cite{han2025regmixmatch}. Since JEPAMatch emphasizes representation geometry, a self-supervised learning paradigm, we also compared it against methods that explicitly merge self-supervised learning with semi-supervised classification. These include CrMatch \cite{fan2023revisiting}, SimMatch \cite{zheng2022simmatch}, and Suave \cite{fini2023semi}.

\textbf{Training Hyperparameters.} To ensure fairness, our codebase is built on the Unified Semi-Supervised Learning Benchmark (USB) \cite{wang2022usb}. The code for implementing JEPAMatch is available at \textcolor{blue}{\url{https://github.com/aah94/JEPAMatch}}. We retained the default training hyperparameters (e.g., learning rate, optimizer settings, and batch sizes) from the USB repository for all baseline methods, with minor adjustments for JEPAMatch. We trained our model for \( 2^{17} \) iterations on CIFAR-100 and STL-10, and \( 2^{18} \) iterations on Tiny-ImageNet. The number of warm-up iterations $T_{\text{warm}}$ can be considered a hyperparameter, which is set to one-third of the total iterations for high-label scenarios and half for low-label scenarios. For loss hyperparameters, we set \( \lambda_{\text{unsup}} = 1 \) and \( \lambda_{\text{rep}} = 0.5 \) across all datasets. The parameter \( \beta \) is set to 0.2 for CIFAR-100 and STL-10, and adjusted to 0.05 for Tiny-ImageNet.

\section{Experimental Results}

\subsection{Main Results}

\paragraph{\textbf{Image Classification.}}

Table~\ref{tab:main_results} presents the error rates (\%) of JEPAMatch and state-of-the-art SSL methods on the CIFAR-100 and STL-10 benchmarks across various labeled data scenarios. The results highlight that JEPAMatch consistently outperforms or remains competitive with existing methods while requiring 8 times fewer iterations ($2^{17}$ vs.\ $2^{20}$) for CIFAR-100 and STL-10, demonstrating its computational efficiency. Notably, JEPAMatch achieves the lowest error rates for CIFAR-100 with 400 (34.25\%) and 2,500 (22.59\%) labeled samples, and it is comparable to the best-performing method for 10,000 labeled samples (18.55\%). On STL-10, JEPAMatch dominates in the 1,000 labeled samples scenario (4.28\%) and remains highly competitive with 40 labeled samples (13.44\%). This efficiency and performance advantage stem from JEPAMatch's ability to leverage all unlabeled data from the first iteration, enabling faster and more stable convergence compared to methods that gradually incorporate unlabeled data. Overall, JEPAMatch sets a new standard in SSL by combining state-of-the-art accuracy with significantly reduced training time.

\begin{table}[ht]
    \centering
    \caption{Error rates (\%) on CIFAR-100 and STL-10 (3 seeds). Best results in \textbf{bold}. $^*$ Standard deviation is not reported in the original paper. -- Results not reported in the original paper.}
    \label{tab:main_results}
    
    \scriptsize 
    \setlength{\tabcolsep}{1.5pt} 
    
    \resizebox{\columnwidth}{!}{%
    \begin{tabular}{l c c c c c c}
        \toprule
        \multirow{2}{*}{Method} & \multirow{2}{*}{Iter.} & \multicolumn{3}{c}{CIFAR-100} & \multicolumn{2}{c}{STL-10} \\
        \cmidrule(lr){3-5} \cmidrule(lr){6-7}
        & & 400 & 2.5K & 10K & 40 & 1K \\
        \midrule
        PseudoLabel \cite{lee2013pseudo}       & $2^{20}$ & $87.45 \pm 0.85$ & $57.74 \pm 0.28$ & $36.55 \pm 0.24$ & $74.68 \pm 0.99$ & $32.64 \pm 0.71$ \\
        MeanTeacher \cite{tarvainen2017mean}   & $2^{20}$ & $81.11 \pm 1.44$ & $45.17 \pm 1.06$ & $31.75 \pm 0.23$ & $71.72 \pm 1.45$ & $33.90 \pm 1.37$ \\
        MixMatch \cite{berthelot2019mixmatch}  & $2^{20}$ & $67.59 \pm 0.66$ & $39.76 \pm 0.48$ & $27.78 \pm 0.29$ & $54.93 \pm 0.96$ & $21.70 \pm 0.68$ \\
        ReMixMatch \cite{berthelot2019remixmatch}& $2^{20}$ & $42.75 \pm 1.05$ & $26.03 \pm 0.35$ & $20.02 \pm 0.27$ & $32.12 \pm 6.24$ & $6.74 \pm 0.14$ \\
        UDA \cite{xie2020unsupervised}         & $2^{20}$ & $46.39 \pm 1.59$ & $27.73 \pm 0.21$ & $22.49 \pm 0.23$ & $37.42 \pm 8.44$ & $6.64 \pm 0.17$ \\
        FixMatch \cite{sohn2020fixmatch}       & $2^{20}$ & $46.42 \pm 0.82$ & $28.03 \pm 0.16$ & $22.20 \pm 0.12$ & $35.97 \pm 4.14$ & $6.25 \pm 0.33$ \\
        FlexMatch \cite{zhang2021flexmatch}    & $2^{20}$ & $39.94 \pm 1.62$ & $26.49 \pm 0.20$ & $21.90 \pm 0.15$ & $29.15 \pm 4.16$ & $5.77 \pm 0.18$ \\
        FreeMatch \cite{wang2022freematch}     & $2^{20}$ & $37.98 \pm 0.42$ & $26.47 \pm 0.20$ & $21.68 \pm 0.03$ & $15.56 \pm 0.55$ & $5.63 \pm 0.15$ \\
        CrMatch \cite{fan2023revisiting}          & $2^{20}$ & $39.45 \pm 1.69$ & $25.43 \pm 0.14$ & $20.40 \pm 0.08$ & -- & $4.89 \pm 0.17$ \\
        SoftMatch \cite{chen2023softmatch}     & $2^{20}$ & $37.10 \pm 0.77$ & $26.66 \pm 0.25$ & $22.03 \pm 0.03$ & $21.42 \pm 3.48$ & $5.73 \pm 0.24$ \\
        SimMatch \cite{zheng2022simmatch}      & $2^{20}$ & $37.81 \pm 2.21$ & $25.07 \pm 0.32$ & $20.58 \pm 0.11$ & -- & -- \\
        FlatMatch \cite{huang2023flatmatch}      & $2^{20}$ & $38.76 \pm 1.62$ & $25.38 \pm 0.85$ & $19.01 \pm 0.43$ & $16.20 \pm 4.34$ & $4.82 \pm 1.21$ \\
        Suave$^{*}$ \cite{fini2023semi} & $2^{20}$ & $35.4$ & $23.00$ & $\mathbf{18.4}$ & -- & -- \\
        RegMixMatch$^{*}$ \cite{han2025regmixmatch}& $2^{20}$ & $35.27$ & $23.78$ & $19.41$ & $\mathbf{11.74}$ & $4.66$ \\
        \midrule
        \textbf{JEPAMatch (Ours)}              & $\mathbf{2^{17}}$ & $\mathbf{34.25 \pm 1.97}$ & $\mathbf{22.59 \pm 1.17}$ & $18.55 \pm 0.85$ & $13.44 \pm 3.2$ & $\mathbf{4.28 \pm 1.43}$ \\
        \bottomrule
    \end{tabular}%
    }
\end{table}

We extended our evaluation to the Tiny-ImageNet dataset by testing JEPAMatch's scalability on complex image distributions. As shown in Table~\ref{tab:tiny_results}, JEPAMatch achieves the highest performance, reaching an error rate of 38.82\% with 1k labeled images split and 24.5\% with 10k labeled images split. These results improve upon baseline methods such as FlexMatch and SoftMatch, confirming that our adaptive geometric representation approach scales effectively to larger datasets with more classes.

\begin{table}[h!]
    \centering
    \caption{Error rates (\%) on Tiny-ImageNet. 
    All methods use the same seed. Best results in \textbf{bold}.}
    \label{tab:tiny_results}
    \begin{tabular}{l c c c}
        \toprule
        \multirow{2}{*}{Method} & \multirow{2}{*}{Iteration} & \multicolumn{2}{c}{Tiny-ImageNet} \\
        \cmidrule(lr){3-4}
        & & 1k & 10k \\
        \midrule
        FlexMatch & $2^{18}$ & 41.73 & 27.89 \\
        SoftMatch & $2^{18}$ & 40.09 & 25.92 \\
        \midrule
        \textbf{JEPAMatch} & $2^{18}$ & $\mathbf{38.82}$ & $\mathbf{24.50}$ \\
        \bottomrule
    \end{tabular}
\end{table}

Table~\ref{tab:baselines}  presents the performance of various approaches on the CIFAR-100 benchmark using 4 labeled examples per class. On the left hand side of Table~\ref{tab:baselines}, JEPAMatch is compared with other baselines under different class imbalance ratios ($\gamma = 20, 50, 100$). The results show that JEPAMatch consistently outperforms baseline methods such as FixMatch, FlexMatch, and SoftMatch, achieving the lowest error rates of 46.27\% ($\gamma=20$), 55.16\% ($\gamma=50$), and 59.93\% ($\gamma=100$).

This performance improvement is attributed to the SIGReg loss, which mitigates class dominance during the warm-up phase by preventing over-representation of majority classes. Subsequently, the adaptive class-wise SIGReg mechanism dynamically allocates representation space to each class, effectively reducing bias toward head classes (those with more data). This allows JEPAMatch to achieve a 1.0--2.0\% improvement over other approaches across all imbalance scenarios.

\begin{table}[t!]          
\caption{ Error rates (\%) on CIFAR-100 with 4 labeled examples per class of FixMatch, FlexMatch and SoftMatch \textbf{(Left)}  with JepaMatch and varying imbalance factors ($\gamma$). \textbf{(Right)} JEPAMatch variants applied to different confidence-based pseudo-labeling methods. Baseline results are taken from \cite{wang2022usb} using WRN-28-2, which differs from the backbone used in Table~\ref{tab:main_results}.}
    \label{tab:baselines}
    \begin{minipage}{0.5\textwidth}
     \begin{adjustbox}{width=\linewidth, center}
\begin{tabular}{l c c c}
                \toprule
                \multirow{2}{*}{Method} & \multicolumn{3}{c}{CIFAR-100 (Imbalance $\gamma$)} \\
                \cmidrule(lr){2-4}
                & $\gamma$ = 20 & $\gamma$ = 50 & $\gamma$ = 100 \\
                \midrule
                FixMatch  & $50.42 \pm 0.78$ & $57.89 \pm 0.33$ & $62.40 \pm 0.48$ \\
                FlexMatch & $49.11 \pm 0.60$ & $57.20 \pm 0.39$ & $62.70 \pm 0.47$ \\
                SoftMatch & $48.09 \pm 0.55$ & $56.24 \pm 0.51$ & $61.08 \pm 0.81$ \\
                \midrule
                \textbf{JEPAMatch} &  $\mathbf{46.27 \pm 0.94}$ & $\mathbf{55.16 \pm 1.12}$ & $\mathbf{59.93 \pm 0.68}$ \\
                \bottomrule
            \end{tabular}
             \end{adjustbox}
                \end{minipage}
    \hfill
    \begin{minipage}{0.48\textwidth}
         \begin{adjustbox}{width=.9\linewidth, center}
              \begin{tabular}{l c c}
            \toprule
            \multirow{2}{*}{Method} & \multirow{2}{*}{Iteration} & CIFAR-100 \\
            \cmidrule(lr){3-3}
            & & 400 \\
            \midrule
            FlexMatch & $2^{20}$ & $50.15 \pm 1.51$ \\
            FreeMatch & $2^{20}$ & $49.64 \pm 1.46$ \\
            SoftMatch & $2^{20}$ & $49.24 \pm 2.16$ \\
            \midrule
            \textbf{JEPAMatch}(Flex)    & $2^{17}$ & $\mathbf{45.77 \pm 2.77}$ \\
            \textbf{JEPAMatch}(Free)    & $2^{17}$ & $\mathbf{45.12 \pm 1.98}$ \\
            \textbf{JEPAMatch}(Soft)    & $2^{17}$ & $\mathbf{44.65 \pm 2.14}$ \\
            \bottomrule
        \end{tabular}   
                     \end{adjustbox}
                \end{minipage}
\end{table}

Beyond its integration with FlexMatch, the right side of Table~\ref{tab:baselines} tests JEPAMatch as a general module by adding it on top of three different confidence-based methods such as FlexMatch, FreeMatch, and SoftMatch on CIFAR-100 with 400 labels. In all three cases, adding our representation level clearly lowers the error rate, giving an absolute improvement for every variant. Moreover, these gains are achieved with $2^{17}$ instead of $2^{20}$ iterations, a massive reduction in training cost. Since the improvement is consistent across all three thresholding schemes, the benefit of JEPAMatch does not depend on the specific pseudo-label selection rule.

\subsection{Empirical Analysis}

\paragraph{\textbf{Convergence Speed.}}

\begin{figure}[htbp]
    \centering
    \includegraphics[width=0.57\linewidth]{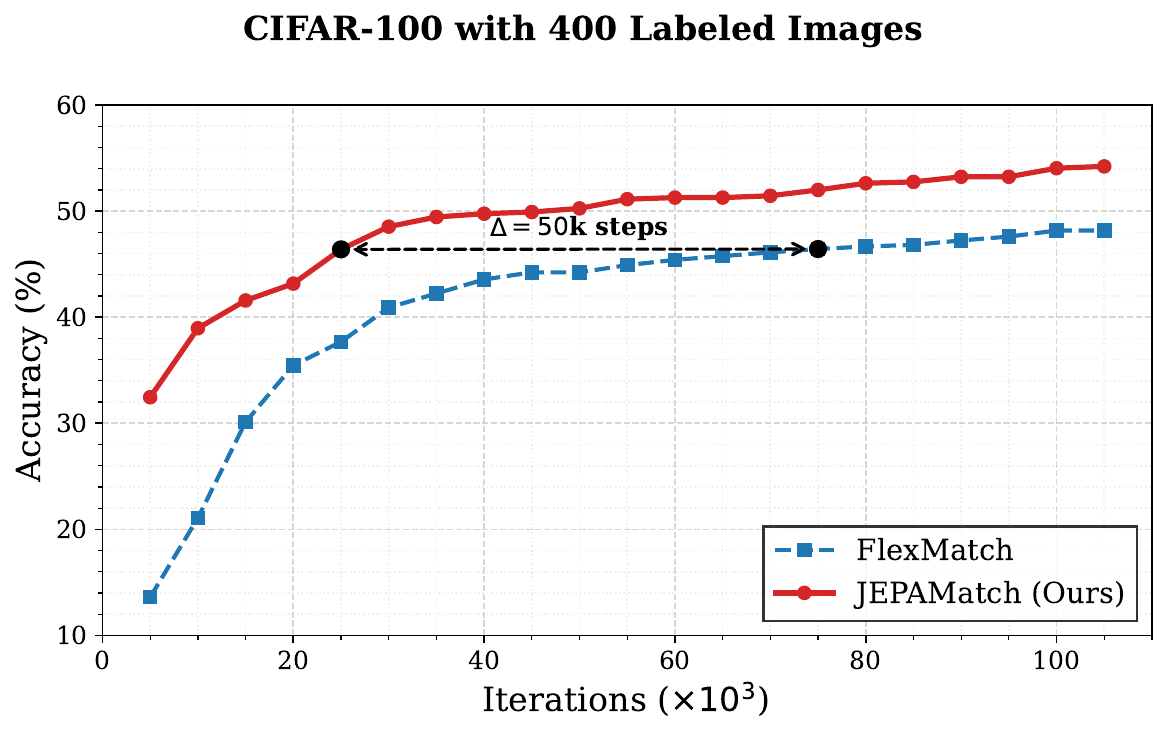}
    
    \caption{Convergence speed and evolution of accuracy of JEPAMatch and FlexMatch over training iterations.}
    \label{fig:convergence_speed}
\end{figure}

Figure \ref{fig:convergence_speed} compares the convergence speed and accuracy of JEPAMatch (red line) and FlexMatch (blue dashed line) on the CIFAR-100 dataset with only 4 labeled samples per class. JEPAMatch demonstrates a clear advantage in both speed and performance: it reaches $46.4\%$ accuracy roughly 50,000 iterations earlier than FlexMatch in terms of training steps. This rapid convergence is particularly valuable in resource-limited settings, as it significantly reduces training time and computational costs. Beyond speed, JEPAMatch also achieves a higher final accuracy ($\sim 54\%$) compared to FlexMatch ($\sim 48\%$), indicating not just faster learning but also better overall model performance. The efficiency of JEPAMatch is further highlighted by its steep accuracy improvement in the early iterations, quickly surpassing FlexMatch and maintaining a consistent lead throughout training.

It is worth to mention that the efficiency advantage of JEPAMatch is 
measured in terms of training iterations rather than wall-clock time 
per iteration, as JEPAMatch introduces additional computational 
components. Nevertheless, the $8\times$ reduction in the number 
of training steps (from $2^{20}$ iterations down to only $2^{17}$) more than compensates for this per-iteration overhead, resulting in a net reduction of overall computational cost. Since JEPAMatch reaches state-of-the-art performance well before the conventional training budget is exhausted, we do not extend training to $2^{20}$ iterations.

\begin{figure}[ht]
    \centering
    
    \begin{subfigure}[b]{0.48\textwidth}
        \centering
        \includegraphics[width=\linewidth]{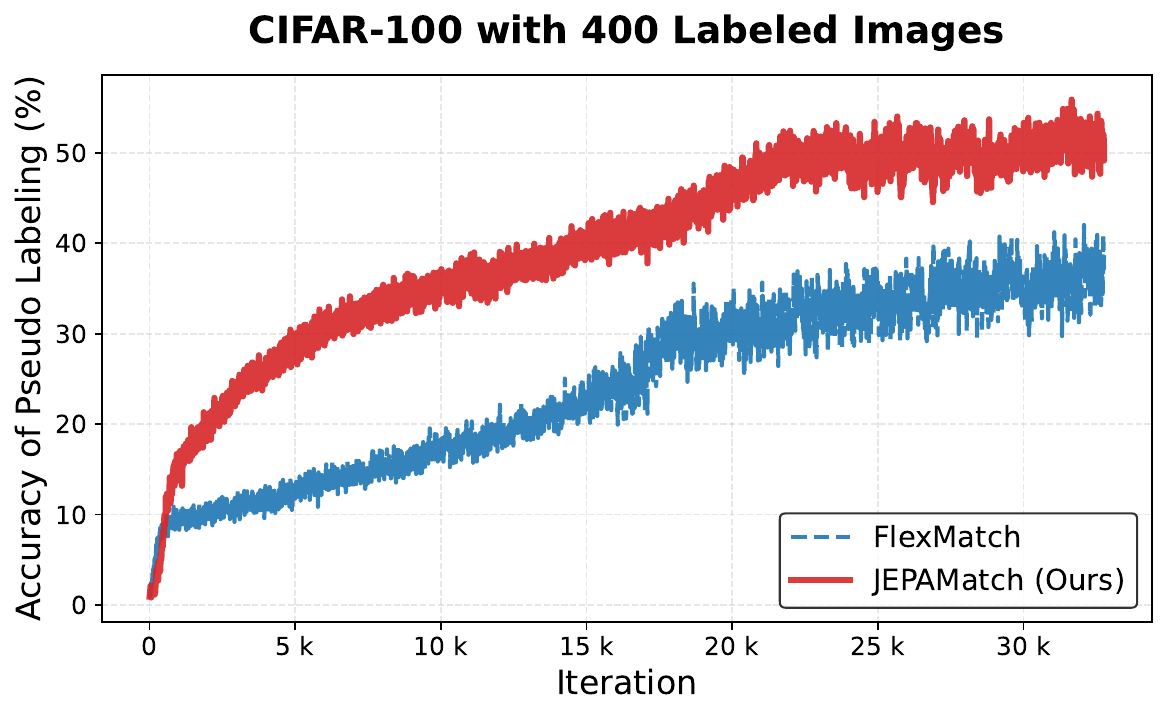}
        \caption{CIFAR-100}
        \label{fig:pla_left}
    \end{subfigure}
    \hfill 
    \begin{subfigure}[b]{0.48\textwidth}
        \centering
        \includegraphics[width=\linewidth]{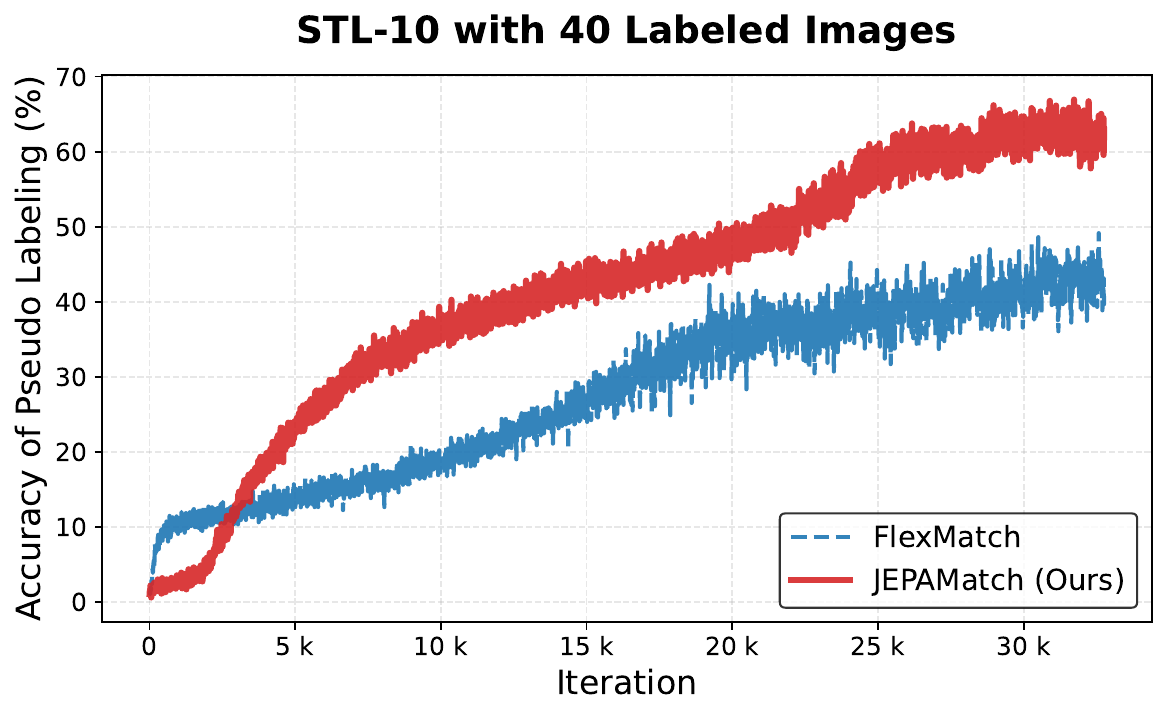}
        \caption{STL-10}
        \label{fig:pla_right}
    \end{subfigure}
    
    \caption{Comparison of pseudo-labeling accuracy between JEPAMatch and FlexMatch over training iterations for CIFAR-100 and STL-10 using 4 labeled examples per class. } 
    \label{fig:pla}
    
\end{figure}

\paragraph{\textbf{Accuracy of Pseudo Labeling.}}
Figure \ref{fig:pla} compares the percentage of true labels in the assigned  pseudo-labels (called pseudo-labeling accuracy) of JEPAMatch (red) and FlexMatch (dashed blue) over training iterations for CIFAR-100 (400 labeled images) and STL-10 (40 labeled images). JEPAMatch consistently achieves higher pseudo-labeling accuracy throughout training, reaching around 50\% for CIFAR-100 and 65\% for STL-10, compared to FlexMatch’s $\sim 35\%$ and $\sim 45\%$, respectively. This superior performance is evident from the early stages, where JEPAMatch converges faster and maintains a stable, upward trajectory, while FlexMatch exhibits more fluctuations and plateaus earlier. The smoother and more consistent accuracy curve of JEPAMatch suggests it is better at filtering noisy pseudo-labels and providing reliable supervision for unlabeled data.
The advantage of JEPAMatch stems from its curriculum-based pseudo-labeling, which dynamically selects high-confidence labels, and its adaptive class-wise SIGReg mechanism, which mitigates class imbalance and reduces bias toward majority classes. Additionally, the use of Joint-Embedding Predictive Architecture  aligns global and local representations, further improving pseudo-label quality.

\begin{figure}[ht]
    \centering
    \includegraphics[width=0.9\linewidth]{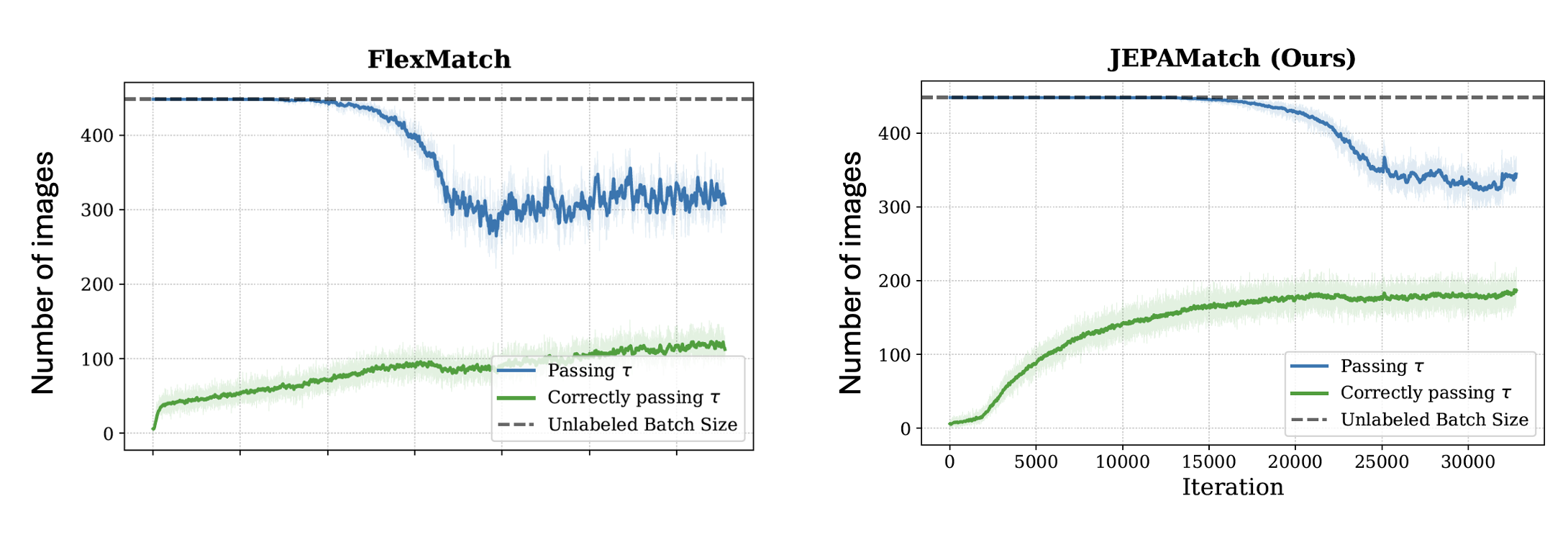}
    \caption{Data Utilization in Pseudo Labeling on CIFAR-100 with 400 labeled images.}
    \label{fig:data_utilization}
    \vspace{-0.3cm}
\end{figure}

\paragraph{\textbf{Data Utilization.}}
\label{app:data_util}
Figure \ref{fig:data_utilization} compares FlexMatch and JEPAMatch on CIFAR-100 in terms of data utilization. For a batch of unlabeled samples, we report (i) in blue, the number of samples that actually pass the confidence threshold and are used for training, and (ii) in green, the number of samples that should pass the threshold according to their true labels. JEPAMatch consistently maintains more pseudo-labels passing the threshold while also achieving higher correctness, allowing the network to leverage a larger portion of unlabeled samples for longer during training. This results in a stronger supervisory signal from unlabeled samples, which in turn improves both unlabeled and overall performances.

\begin{figure}[ht]
    \centering
    \vspace{-1cm}
    \includegraphics[width=\linewidth]{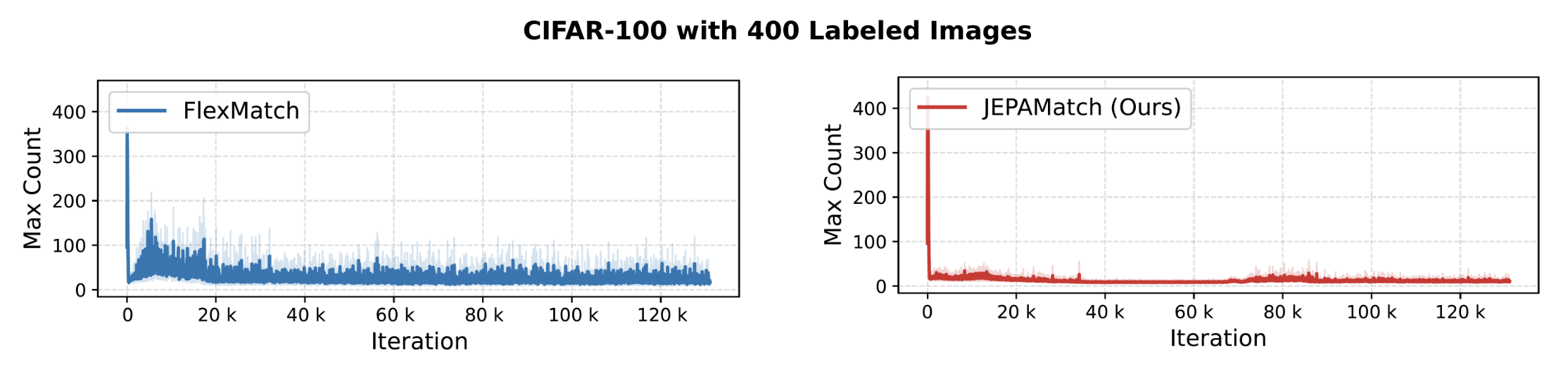}
    \caption{\textbf{Maximum class count} evolution.}
    \label{fig:max_count}
    \vspace{-0.3cm}
\end{figure}

For a batch of unlabeled samples, Figure \ref{fig:max_count} shows the number of pseudo-labels assigned to the predominant class (i.e., the class receiving the highest number of pseudo-labels) for both FlexMatch and JEPAMatch on CIFAR-100. In FlexMatch, the major class can receive up to one quarter of all pseudo-labels, indicating strong class dominance. In contrast, our method maintains a stable maximum of around 30 pseudo-labels for the major class during training. This demonstrates that JEPAMatch significantly reduces class dominance. 

\begin{figure}[ht]
    \centering

    \includegraphics[width=0.8\linewidth]{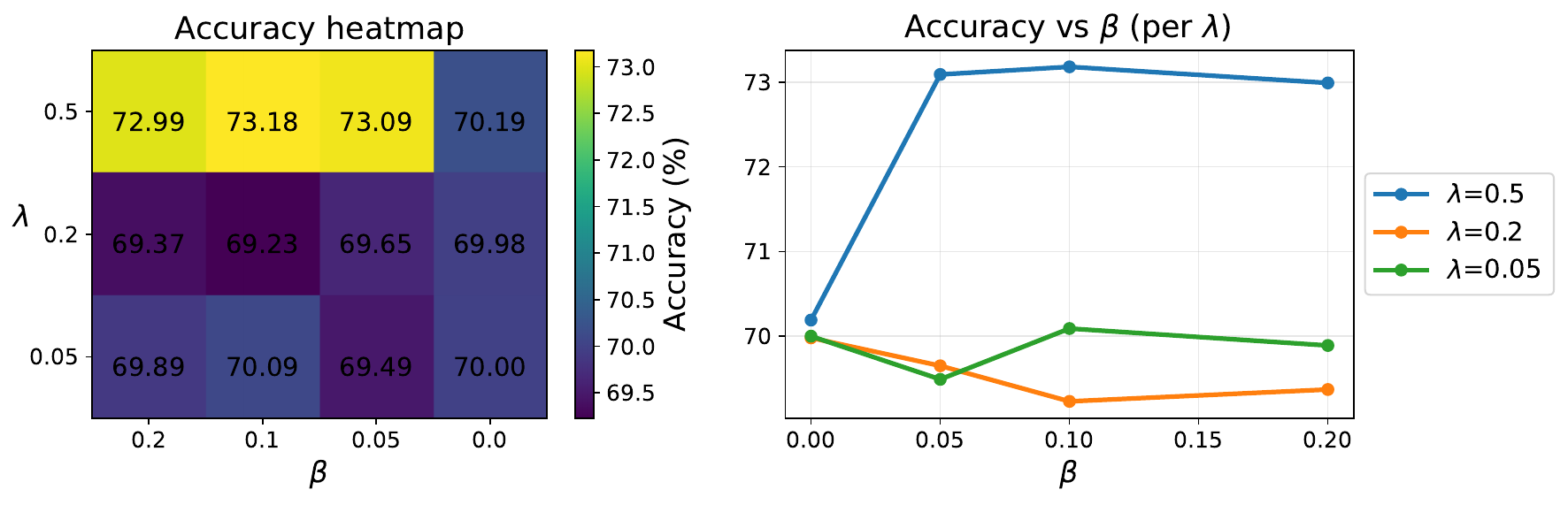}
    \caption{Effect of different hyper-parameters $\lambda_{rep}$ and $\beta$ on CIFAR-100 with 10000 labeled images. The best performance is achieved with $\lambda_{rep}=0.5$ and $\beta=0.1$.}
    \label{fig:Hyper}
\end{figure}

\paragraph{\textbf{Hyper-parameters.}}
\label{app:hyper}
To understand the impact of our loss hyper-parameters, we evaluate the representation weight $\lambda_{\text{rep}}$ (denoted here as $\lambda$) and the SIGReg balance parameter $\beta$. As shown in Figure \ref{fig:Hyper}, the accuracy heatmap reveals the following trends. 
The model achieves its best accuracy of 73.18\% when configured with $\lambda = 0.5$ and $\beta = 0.1$. Reducing the SIGReg regularization ($\beta = 0.0$) at this optimal $\lambda$ leads to a sharp drop in performance to 70.19\%. 

Conversely, lowering the representation weight restricts the model to a lower accuracy range. For instance, with $\lambda = 0.2$, the accuracy varies between 69.23\% and 69.98\% , while $\lambda = 0.05$ results in accuracies between 69.49\% and 70.09\%. Overall, these results confirm that a carefully balanced contribution of both the predictive representation loss ($\lambda$) and the geometric regularization ($\beta$) is crucial for achieving optimal performance.

\section{Conclusion}

In this work, we introduced JEPAMatch, a novel semi-supervised learning framework that integrates the theoretically grounded architecture of LeJEPA with the adaptive thresholding capabilities of FlexMatch. By addressing the inherent limitations of confidence-based pseudo-labeling, JEPAMatch decouples the discrete classification task from geometric organization of the representation space. This decoupling is achieved through a two-level learning process: Curriculum Level for pseudo-label selection and a Representation Level for structuring the feature space. The joint optimization of these levels not only enhances classification performance but also mitigates the class imbalance issues commonly associated with confidence-based methods, leading to faster and more stable convergence.
Our empirical results, including the convergence analysis on different benchmarks demonstrate that JEPAMatch improves accuracy while requiring substantially fewer optimization steps to reach strong performance. It achieves higher final accuracy while converging significantly faster, making it a more robust and resource-efficient solution for semi-supervised learning tasks. This work paves the way for future research in leveraging structured representation learning for semi-supervised learning.

%
%
%
\bibliographystyle{splncs04}
\bibliography{mybibliography}

\end{document}